\documentclass{article}

\usepackage[final]{nips_2016}

\usepackage[utf8]{inputenc} \usepackage[T1]{fontenc}    \usepackage[hidelinks]{hyperref}       \usepackage{url}            \usepackage{booktabs}       \usepackage{amsfonts}       \usepackage{nicefrac}       \usepackage{microtype}
\usepackage{amssymb,amsmath}
\usepackage{amsthm}
\usepackage{bm}
\def\given{\;\middle\vert\;}

\def\indicator{\delta}

\DeclareMathOperator*{\expectation}{\mathbb{E}}
\def\prob{P}

\def\defeq{\dot=}

\newtheorem{theorem}{Theorem}

\newtheorem{corollary}{Corollary}
\newcommand{\mat}[1]{#1}\def\eye{\mat{I}}

\def\state{s}
\def\states{\mathcal{S}}
\def\action{a}

\def\actions{\mathcal{A}}
\def\option{w}
\def\opt{\option}

\def\options{\mathcal{W}}

\def\inv{{-1}}
 \usepackage{mathtools}
\usepackage{mathabx}
\usepackage{dsfont}

\title{A Matrix Splitting Perspective on Planning with Options}
\author{
  Pierre-Luc Bacon and Doina Precup \\
  McGill University\\
  \texttt{\{pbacon, dprecup\}@cs.mcgill.ca}
}

\begin{document}
\maketitle
\begin{abstract}
  We show that the Bellman operator underlying the options framework leads to
  a matrix splitting, an approach traditionally used to speed up convergence of
  iterative solvers for large linear systems of equations. Based on standard
  comparison theorems for matrix splittings, we then show how the asymptotic rate of
  convergence varies as a function of the inherent timescales of the options.
  This new perspective highlights a trade-off between asymptotic performance
  and the cost of computation associated with building a good set of options.
\end{abstract}

\section{Introduction}

The ability to represent and plan with temporally extended actions has long been
recognized as an essential component for scaling up reinforcement learning systems.
The \textit{options} framework \citep{Sutton1999opt, Precup2000} formalized this idea
and showed how temporally extended actions can be used for learning and planning
in reinforcement learning. Options can sometimes lead to better and faster
exploration, learning, planning and transfer \citep{Sutton1999opt} while being robust to
model misspecifications and uncertainty \citep{He2011}.
While efficient algorithms for learning options have recently been proposed
\citep{Bacon2016, Mankowitz2016, Vezhnevets2016, Daniel2016},
there is still no consensus on what constitute \textit{good options}.

In this paper, we show that a choice of options is equivalent to a choice
of iterative algorithm for solving Markov decision problems. We reach this
conclusion by noting that the generalized Bellman operator underlying
options and their models admits a linear representation as a
\textit{matrix splitting} \citep{Varga1962, Young1971, Puterman1994},
a notion which comes in pair with that of matrix preconditioning. As with options,
the goal of these methods is to transform a linear system into one with
the same solution but which is easier to solve. With this new perspective, the
answer to what good options are becomes clearer while offering new theoretical
tools to analyze them.

\section{Background and Notation}

We restrict our attention to the class of discounted Markovian decision problems
with finite state and action spaces.
A discounted Markov Decision Process is specified by a finite set of states $\states$,
a finite set of actions $\actions$, a reward function $r: \states \times \actions \to \mathbb{R}$,
a transition function $\prob :\states \times \actions \to (\states \to [0,1])$
and a discount factor $\gamma \in [0, 1)$. We write $\pi\left(a \given s\right)$
to denote the probability of taking action $a$ in state $s$ under the stochastic
policy $\pi$: that is $\sum_{a \in \actions} \pi\left(a \given s\right) = 1,\, \forall s \in \states$.
The value function $v_\pi : \states \to \mathbb{R}$ represents the expected sum of discounted
rewards encountered along the trajectories induced by the MDP and policy:
$v_\pi(s) \defeq \expectation\left[ \sum_{k=0}\gamma^k r(S_k, A_k) \given S_0 = s, \pi\right]$.
We also write $r_\pi(s) \defeq \sum_{a \in \actions} \pi\left( a \given s\right) r(s, a)$
and $\prob_\pi(s, s') \defeq  \sum_{a \in \actions} \pi\left( a \given s\right) \prob\left(s' \given s, a\right)$.
The spectral radius of a $n \times n$
matrix $A$ with eigenvalues $\lambda_i, i \in [0, n]$ is $\rho(A) \defeq \max_{1 \leq i \leq n} | \lambda_i |$.

An option $\opt \in \options$ is a triple $(\mathcal{I}_\opt, \pi_\opt, \beta_\opt)$
where $\mathcal{I}_\opt \subseteq \states$ is the initiation set of $\opt$,
$\pi_\opt$ is its policy, and $\beta_\opt: \states \to [0, 1]$
is its termination function. In addition, there is
a policy over options $\mu : \states \to (\options \to [0, 1])$ whose role
is to select an option whenever a termination event is sampled from the termination
function of the current option. In the call-and-return model of execution,
$\mu$ picks among $\{ \opt \in \options : s \in \mathcal{I}_\opt \}$ in
state $s$ and executes the policy of the selected option irrevocably
until termination. We distinguish the call-and-return model from what we call
\textit{gating execution}, in which the choice of option is reconsidered at every step.
The results in sections \ref{sect:genbell-gating} and \ref{sect:matrixsplitting}
apply only to the gating model. However, we also show
in section \ref{sect:callreturn} that the call-and-return execution model can be studied
in the matrix splitting framework. For simplicity, we finally assume that options
are available everywhere, that is :
$\mathcal{I}_\opt = \states,\, \forall \opt \in \options$.

\section{Generalized Bellman Equations for Gating Execution}
\label{sect:genbell-gating}
Planning with the value iteration algorithm over primitive actions usually involves a Bellman operator
$T$ of the form: $(Tv)(s) = r_\pi(s) + \gamma \sum_{s' \in \states} \prob_\pi(s, s') v(s')$.
Rather than backing up values for only one step ahead, \cite{Sutton1995}
showed that multi-steps backups can equally be used for planning as long
as the corresponding generalized Bellman operators satisfy the
Bellman equations. We can extend this idea  \citep{Precup1998} by making the number of Bellman
backups a random variable which is determined by the termination events of an
options-based process. Let $K$ be a random variable representing the number of backups
performed per iteration, we then define our generalized Bellman operator as follows:
\begin{align}
  (L v)(s) \defeq \expectation\left[ \sum_{k=0}^{K-1} \gamma^k r(S_k, A_k) + \gamma^K v(S_K)\given S_0 = s, \options, \mu, v \right] \enspace . \label{eq:genbell}
\end{align}
By linearity and with Markov options, we can decompose $L$ into a
reward and a transition part. The \textit{reward model}
$b: \states \to \mathbb{R}$ underlying $L$ is the discounted sum of rewards
until termination, averaged over all options:
\begin{align}
 b(s) &\defeq \expectation\left[ \sum_{k=0}^{K-1} \gamma^k r(S_k, A_k) \given S_0 = s, v \right]
     = r_\sigma(s) + \gamma \sum_{s'} P_\sharp(s, s') b(s') \label{eq:brec}\enspace ,
 \end{align}
where $\sigma$ and $P_\sharp$ are defined as follows:
\begin{align*}
 \sigma\left(a \given s\right) &\defeq \sum_{\opt} \mu\left( \opt \given s \right) \pi_\opt\left( a \given s \right),\enspace
  P_\sharp(s, s')  \defeq \sum_{\opt} \mu\left(\opt \given s\right) \sum_{a} \pi_\opt\left( a \given s\right) P\left(s' \given s, a\right)(1 - \beta_\opt(s')) \enspace .
 \end{align*}
The \textit{sharp} $\sharp$ symbol here stands for ``continuation''.
Likewise, the \textit{transition model} $F:\states \to (\states \to [0, 1])$ has
the following recursive form:
\begin{align}
  F(s, s') &\defeq \gamma P_\bot(s, s') + \gamma \sum_{\bar{s}} P_\sharp(s, \bar{s})F(\bar{s}, s') \label{eq:Frec}\enspace .
\end{align}
where $P_\bot(s, s') \defeq \sum_{\opt} \mu\left(\opt \given s\right) \sum_{a} \pi_\opt\left( a \given s\right) P\left(s' \given s, a\right)\beta_\opt(s')$
and $\bot$ stands for ``termination''.
The linear system  of equations corresponding to the reward and transition models
admits a solution given that the matrix $\eye - \gamma P_\sharp$ is nonsingular (which we prove below):
\begin{align}
  b = (\eye - \gamma P_\sharp)^\inv r_\sigma,\enspace F =  (\eye - \gamma P_\sharp)^\inv (\gamma P_\bot) \enspace . \label{eq:bF}
\end{align}

The generalized Bellman operator $L$ then becomes:
\begin{align}
  Lv &= b + Fv = (\eye - \gamma P_\sharp)^{-1} r_\sigma + \gamma(\eye - \gamma P_\sharp)^{-1} P_\bot v \label{eq:LbFv}\enspace .
\end{align}

\cite{VanNunen1976} showed that the basic iterative methods such as the
Gauss-Seidel, Jacobi, Successive Overrelaxation, or Richardson's variants \citep{Puterman1994} of value iteration can be obtained
through different \textit{stopping time} functions in an operator of the same form
as \eqref{eq:genbell}, or equivalently, as a linear \textit{transformation}
of an MDP into an equivalent one. This perspective was leveraged by \cite{Porteus1975} to
derive better bounds by transforming an MDP into one which has the same
optimal policy but whose spectral radius is smaller. The idea that stopping
time functions (termination functions) lead to a transformation
of an MDP is also what we just found by writing \eqref{eq:genbell} as \eqref{eq:LbFv}.
Using Porteus' terminology, \eqref{eq:LbFv} in fact corresponds a
\textit{pre-inverse transformation} through the reward and transition models $b$ and $F$.
The pre-inverse transform as well as the basic iterative methods can also be
studied more generally via the notion of \textit{matrix splittings} \citep{Varga1962}
developed in the context of matrix iterative analysis.

\section{Matrix Splitting}
\label{sect:matrixsplitting}
For $A = M - N \in \mathbb{R}^{n\times n}$ and provided
that  $A$ and $M$ are nonsingular, \cite{Varga1962} showed that the iterates
$x_{k+1} = M^\inv N x_k + M^\inv b$ converge to the unique solution of the linear
system of equation $Ax = b$.
In the policy evaluation problem, we are working with the linear system of
equations $(\eye - \gamma P_\sigma)v = r_\sigma$ where $\sigma$ is the target policy
to evaluate. We now show that the reward
and transition models \eqref{eq:bF} precisely corresponds to the notion of a
matrix splitting for the matrix $\eye - \gamma P_\sigma$.
\begin{theorem}[Matrix Splitting in the Gating Model]
  \label{thm:splitting}
  Let $A \defeq \eye - \gamma P_\sigma$, $M \defeq \eye - \gamma P_\sharp$ and $N \defeq \gamma P_\bot$,
  then $A = M - N$ is a regular splitting. \\
  \textbf{Proof:} $M$ is an ``\textit{M-matrix}'' (see chapter 6 of \cite{Berman1979}).
  \textit{M-matrices} have the property of being \textit{inverse-positive}, that is
  $M^\inv$ exists with $M \geq 0$, fulfilling the definition
  of a regular splitting (see def. 3.5 of \cite{Varga1962})
\end{theorem}

\begin{corollary}
  \label{cor:converge}
      For the regular splitting of theorem \ref{thm:splitting},
  \begin{enumerate}
    \item $\rho(\gamma (\eye - \gamma P_\sharp)^{-1}  P_\bot) < 1$
    \item The successive approximation method based on the generalized
    Bellman operator \eqref{eq:LbFv} converges for any initial vector $v_0$.
  \end{enumerate}
  \textbf{Proof:} This follows directly from the fact that options induce
  a regular splitting through the operator $L$ in the gating model. See \cite[Theorem 3.13]{Varga1962}
  for a general proof.
\end{corollary}

Since a set of options and the policy over them induce a matrix splitting, a choice of
options is in fact a \textit{choice of algorithm} for solving MDPs. An important property
of an iterative solver, besides its computational efficiency, is that it should converge
to a solution of the original problem. We should therefore ask ourselves whether
the iterates corresponding to \eqref{eq:LbFv} converge to the true
value function underlying a given target policy. Theorem \ref{thm:policyEvalConsistency}
shows that the successive approximation method induced by a set of options and
policy over them is \textit{consistent}  \citep{Young1971} given that
the marginal action probabilities is equal to the target policy.

\begin{theorem}[Consistency of Policy Evaluation in the Gating Model]
  \label{thm:policyEvalConsistency}
  The iterative method associated with the splitting \eqref{eq:LbFv}
  \begin{align*}
    v_{k+1} = (\eye - \gamma P_\sharp)^\inv r_\sigma + \gamma (\eye - \gamma P_\sharp)^\inv P_\bot v_k, \enspace k \geq 0
  \end{align*}
  is a consistent policy evaluation method in the gating model if the set of options and policy
  over them is such that
  $\sigma\left(a \given s \right) = \sum_\opt \mu\left( \opt \given s \right) \pi_\opt\left( a \given s \right) \, \forall \action \in \actions, \state \in \states$
  where $\sigma$ is the target policy to be evaluated.\\
  \textbf{Proof:}
  Let $v_{\options, \mu}$ be the unique solution to the generalized Bellman equations
  \eqref{eq:LbFv}, we have:
  \begin{align*}
    v_{\options, \mu} &= \left(\eye - \gamma(\eye - \gamma P_\sharp)^{-1}P_\bot \right)^{-1} (\eye - \gamma P_\sharp)^{-1}r_\sigma \\
    &= \left((\eye - \gamma P_\sharp)^{-1}\left((\eye - \gamma P_\sharp) - \gamma P_\bot \right)\right)^{-1} (\eye - \gamma P_\sharp)^{-1}r_\sigma \\
    &= \left(\eye - \gamma P_\sigma \right)^{-1} r_\sigma = v_\sigma
  \end{align*}
  Therefore $v_{\options, \mu}$ is also the solution to the policy evaluation problem
  for the policy $\sigma$.
\end{theorem}

While many set of options can satisfy the marginal condition in the gating model,
not all of them would converge equally fast. Using
comparison theorems for regular matrix splittings \citep{Varga1962} we can
better understand the effect of modelling the world at different timescales on
the asymptotic performance of the induced algorithms.

\begin{theorem}[Predict further, plan faster]
  \label{thm:comp}
In the gating model, if a set of options $\widetilde{\options}$ has the same intra-option policies
and policy over options with some other set $\options$ but whose termination
functions  are such that
$\beta_{\widetilde\opt}(s) \leq \beta_{\opt}(s)\, \forall \opt \in \options, \state \in \states$,
then $0 \leq \rho(\widetilde{M}^\inv \widetilde N) \leq \rho(M^\inv N)  < 1$.

\textbf{Proof:} By theorem \ref{thm:splitting}, the two sets of options induce
a corresponding regular splitting. The claim then follows
from \citep[Theorem 3.32]{Varga1962} since $\widetilde{N} \leq N$ (componentwise):
\begin{align*}
  \widetilde{N}(s, s')  &\leq  \gamma \sum_\opt \mu\left( \opt \given s\right)\sum_{a} \pi_\opt\left(a \given s\right) \sum_{s'} \prob\left(s' \given s, a \right) \beta_\opt(s') = N \enspace .
\end{align*}
\end{theorem}
Theorem \ref{thm:comp} consolidates the idea that \textit{modelling}
the world over longer time horizons increases the asymptotic rate
of convergence. This also becomes apparent when writing \eqref{eq:LbFv} in the
following form:
\begin{align}
    (\eye - \gamma P_\sharp) v     &= (\eye - \gamma P_\sharp) v + (r_\sigma - (\eye - \gamma P_\sigma)v)\notag \\
v  &= v + (\eye - \gamma P_\sharp)^\inv (r_\sigma - (\eye - \gamma P_\sigma)v) \label{eq:richardson}
\end{align}
Therefore, options enter the linear system of equations $(\eye - \gamma P_\sigma) v = r_\sigma$
through the preconditioning matrix $M$  \citep{Saad2003}  and yield the following
transformed linear system of equations:
\begin{align}
(\eye - \gamma P_\sharp)^{-1} (\eye - \gamma P_\sigma) v = (\eye - \gamma P_\sharp)^{-1} r_\sigma \label{eq:presys}
\end{align}
As the options timescales increase and $\beta_\opt(s) = 0 \, \forall \opt \in \options, \state \in \states$,
then $P_\sharp = P_\sigma$ and the solution is obtained directly on the right hand
side of \eqref{eq:presys}. The corresponding generalized Bellman operator
also becomes $L^{(\infty)} v  \defeq (\eye - \gamma P_\sigma)^{-1} r_\sigma$ and solves
the original system in one iteration. On the other hand, if the termination
functions are such that the options terminate after only one step, we get
$L^{(0)} v \defeq r_\sigma + \gamma P_\sigma v$, the usual one-step Bellman
operator $T$ of the value iteration algorithm.
Since the spectral radius associated with matrix splitting methods is given by
$\rho(M^\inv N)$, we also have the following:
\begin{align*}
  \rho(M^\inv N) = \rho(M^\inv(M - A)) = \rho\left( \eye - (\eye - \gamma P_\sharp)^{-1} (\eye - \gamma P_\sigma) \right) \leq \|\eye - (\eye - \gamma P_\sharp)^{-1} (\eye - \gamma P_\sigma)\|\enspace .
\end{align*}
This suggests that in terms of asymptotic performances,
a good set of options should be such that it induces a preconditioning matrix
$M$ that is \textit{close} to $\eye - \gamma P_\sigma$ in some sense
but whose inverse $M^\inv$ is easier to compute.

\section{Call-and-Return Execution Model}
\label{sect:callreturn}
The gating execution model assumed so far does not account
for the notion of \textit{temporal commitment} provided by the call-and-return
model. Since a choice of option is made at every step in the gating model,
the policy over option can be decoupled from the termination functions. This gives
us the ability to express the value function as the solution of a matrix splitting
over states. However, it is known \citep{Sutton1999opt} that a set of options with call-and-return execution and an
MDP induce a semi-Markov Decision Process (SMDP), even in the case of Markov options. This means that the trajectories
over state and actions generated with options might no longer correspond to the dynamics of a Markov process. Hence, the existence of an equivalent \textit{marginal}
policy $\sigma(a | s)$ in theorem \ref{thm:policyEvalConsistency} cannot be guaranteed under the call-and-return model.

To restore the Markov property with call-and-return execution, the choice of option
must be remembered as part of the state.  The resulting process defines
a Markov chain in an augmented state space over state-option pairs \citep{Bacon2017}.
This conditioning on both states and options is key to the derivation of the Bellman-like expressions
of \cite{Sutton1999opt}  for the reward and transition models of options.
In the following, we show that the solution to these equations also yields a matrix splitting.

\cite{Sutton1999opt} showed that the reward model of an option can be written recursively as:
\begin{align*}
\mat b_\opt(s) = \sum_{a} \pi_\opt\left(a \given s\right) \left[ r(s, a) + \
\gamma \sum_{s'} \prob\left(s' \given s, a\right) \beta_\opt(s') \mat b_\opt(s') \right] \enspace .
\end{align*}

If we define $\mat r_\opt(s) \defeq \sum_{a} \pi_\opt\left(a \given s\right) r(s, a)$ and
$  \mat P_{w, \sharp, }(s, s') \defeq \sum_{a}  \pi_\opt\left(a \given s\right) \prob\left(s' \given s, a\right)(1 - \beta_\opt(s')$,
we can see that the fixed point of these equations is :
\begin{align*}
  \mat b_\opt = \sum_{t=0}^\infty  \left(\gamma \mat P_{w, \sharp}\right)^t \mat r_\opt = \left( \mat \eye - \gamma \mat P_{w,\sharp}\right)^\inv \mat r_\opt \enspace .
\end{align*}
Similarly, the transition model of an option also admits a set of Bellman-like equations
of the form:
\begin{align*}
  \mat F_{w}(s, s') &= \gamma \sum_{a}  \pi_\opt\left(a \given s\right)\sum_{\bar{s}}\prob\left(\bar{s} \given s, a\right)\left[\beta_\opt(\bar{s})\indicator_{\bar{s} = s'} + (1 - \beta_\opt(\bar{s}))\mat F_{w}(\bar{s}, s') \right] \enspace ,
\end{align*}
whose fixed point can be written using the termination operator $\mat P_{w, \bot}(s, s') \defeq \sum_{a}  \pi_\opt\left(a \given s\right) \prob\left(s' \given s, a\right) \beta_\opt(s)$:
\begin{align*}
  \mat F_{w} = \sum_{t=0}^\infty  \left(\gamma \mat P_{w,\sharp}\right)^t (\gamma \mat P_{w,\bot}) = \left( \mat \eye - \gamma \mat P_{w,\sharp}\right)^\inv (\gamma \mat P_{w,\bot}) \enspace .
\end{align*}

Therefore, with $M_\opt \defeq \eye - \gamma P_{\opt, \sharp}$ and $N_\opt \defeq \gamma P_{\opt, \bot}$ we
have $M_\opt - N_\opt = \eye - \gamma P_{\pi_\opt}$ as a splitting.

\section{Implications}

Given the interpretation of options in terms of
matrix splittings, and consequently as preconditioners, it comes as no surprise
that preconditioning methods share the same goals as options.
Indeed in both cases we seek a representation of the problem which is easier to
solve than the original one. As Herbert Simon wrote in his \textit{Sciences of the Artificial}:
``[...] solving a problem simply means representing it so as to make the solution transparent'' \citep{Simon1969}.
Hence, the problem of finding good options or preconditioners is closely
related to the \textit{representation learning} problem \citep{Minsky1961}.

As with options, the design of general and fast preconditioners is a
longstanding problem of numerical analysis. In some cases, good preconditioners
can be found when problem-specific knowledge is available. However, manual design of
preconditioners, and of options, quickly become a tedious process for large
problems or when only partial knowledge about the domain is available. This is
especially true in the context of reinforcement learning where the MDP is
assumed to be unknown or too large to manipulate directly.
When solving a single problem with options, it is also clear from the connection with
preconditioners that the initial setup cost and subsequent cost per preconditioned iteration
should not outweigh the cost associated with the original problem.
This leads to a fundamental tension between the improvement effort per iteration
and the number of overall iterations.
In our framework, this tension exists between the two poles $L^{(0)}$ and $L^{(\infty)}$.
While $L^{(\infty)}$ has the fastest convergence, it also is just as expensive
as solving the original problem directly.  This is a reminder that modelling
\textit{far} comes with a cost: computational here but also statistical in
the learning case. In fact, the choice of timescale for options also falls under
the same bias-variance tradeoff as for the $\lambda$-operator
\citep{Kearns2000} with $L^{(0)}$ mirroring the choice $\lambda = 0$ and $\lambda = 1$ for $L^{(\infty)}$.

Computational expenses associated with building preconditioners
can be amortized throughout related problems. With options, this would corresponds
to the typical case in which options are reused to speed up learning and planning in a transfer
or continual learning setting. Reusability of options can take place for
example when the transition structure remains fixed but the reward function on the
right-hand side of \eqref{eq:presys} changes. The accounting exercise necessary
to justify the use of options in a transfer setting was considered by
\cite{Solway2014}. From a Bayesian perspective, the authors showed that a particular
kind of bottleneck options is in fact \textit{optimal} when a learning system
must solve a series of related tasks. The idea of adapting the options structure
to minimize the computational effort associated with solving a single task was
also explored in \cite{Bacon2015b}. Using the bounded rationality framework,
it was argued that options should primarily help computationally restricted
adaptive systems: an idea which naturally fits with the preconditioning point
of view.

\bibliographystyle{named}
\bibliography{references}

\end{document}